\documentclass[journal]{vgtc}                % final (journal style)
\ifpdf%                                % if we use pdflatex
  \pdfoutput=1\relax                   % create PDFs from pdfLaTeX
  \usepackage{graphicx}                % allow us to embed graphics files
  \DeclareGraphicsExtensions{.pdf,.png,.jpg,.jpeg} % for pdflatex we expect .pdf, .png, or .jpg files
\else%                                 % else we use pure latex
  \usepackage{graphicx}                % allow us to embed graphics files
  \DeclareGraphicsExtensions{.eps}     % for pure latex we expect eps files
\fi%

\graphicspath{{figures/}{pictures/}{images/}{./}} % where to search for the images

\usepackage{microtype}                 % use micro-typography (slightly more compact, better to read)
\PassOptionsToPackage{warn}{textcomp}  % to address font issues with \textrightarrow
\usepackage{textcomp}                  % use better special symbols
\usepackage{mathptmx}                  % use matching math font
\usepackage{times}                     % we use Times as the main font
\usepackage{cite}                      % needed to automatically sort the references
\usepackage{tabu}                      % only used for the table example
\usepackage{booktabs}                  % only used for the table example
\usepackage{color}
\usepackage{float}
\usepackage{dblfloatfix}
\usepackage{wrapfig}
\usepackage{xspace}
\newcommand*{\eg}{e.g.\@\xspace}
\newcommand*{\ie}{i.e.\@\xspace}
\newcommand*{\etal}{et al.\@\xspace}

\onlineid{1037}

\vgtccategory{Research}
\vgtcpapertype{system}

\title{$\textit{Mo}^\textit{2}\textit{Cap}^\textit{2}$: Real-time \textit{Mo}bile 3D \textit{Mo}tion \textit{Cap}ture with a \textit{Cap}-mounted Fisheye Camera}

\author{Weipeng Xu, Avishek Chatterjee, Michael Zollh\"ofer, Helge Rhodin, Pascal Fua, \textit{Fellow, IEEE},\\ Hans-Peter Seidel, and Christian Theobalt, \textit{Member, IEEE}}
\authorfooter{
\item
 Weipeng Xu, Avishek Chatterjee, Hans-Peter Seidel and Christian Theobalt are with the Max-Planck Institute for Informatics. Emails: wxu, achatter, hpseidel, theobalt@mpi-inf.mpg.de.
\item
 Michael Zollh\"ofer is with Stanford University. E-mail: zollhoefer@cs.stanford.edu.
\item
 Helge Rhodin and Pascal Fua are with the CVLAB of EPFL. E-mails: helge.rhodin, pascal.fua@epfl.ch.
}

\shortauthortitle{xu \MakeLowercase{\textit{et al.}}: $\textit{Mo}^\textit{2}\textit{Cap}^\textit{2}$: Real-time \textit{Mo}bile 3D \textit{Mo}tion \textit{Cap}ture with a \textit{Cap}-mounted Fisheye Camera}
\abstract{
We propose the first real-time system for the egocentric estimation of 3D human body pose in a wide range of unconstrained everyday activities. This setting has a unique set of challenges, such as mobility of the hardware setup, and robustness to long capture sessions with fast recovery from tracking failures. We tackle these challenges based on a novel lightweight setup that converts a standard baseball cap to a device for high-quality pose estimation based on a single cap-mounted fisheye camera. From the captured egocentric live stream, our CNN based 3D pose estimation approach runs at 60~Hz on a consumer-level GPU. In addition to the lightweight hardware setup, our other main contributions are: 1) a large ground truth training corpus of top-down fisheye images and 2) a disentangled 3D pose estimation approach that takes the unique properties of the egocentric viewpoint into account. As shown by our evaluation, we achieve lower 3D joint error as well as better 2D overlay than the existing baselines.
} % end of abstract


\CCScatlist{ % not used in journal version
 \CCScat{K.6.1}{Management of Computing and Information Systems}%
{Project and People Management}{Life Cycle};
 \CCScat{K.7.m}{The Computing Profession}{Miscellaneous}{Ethics}
}

\teaser{
  \centering
  \includegraphics[width=0.7\linewidth]{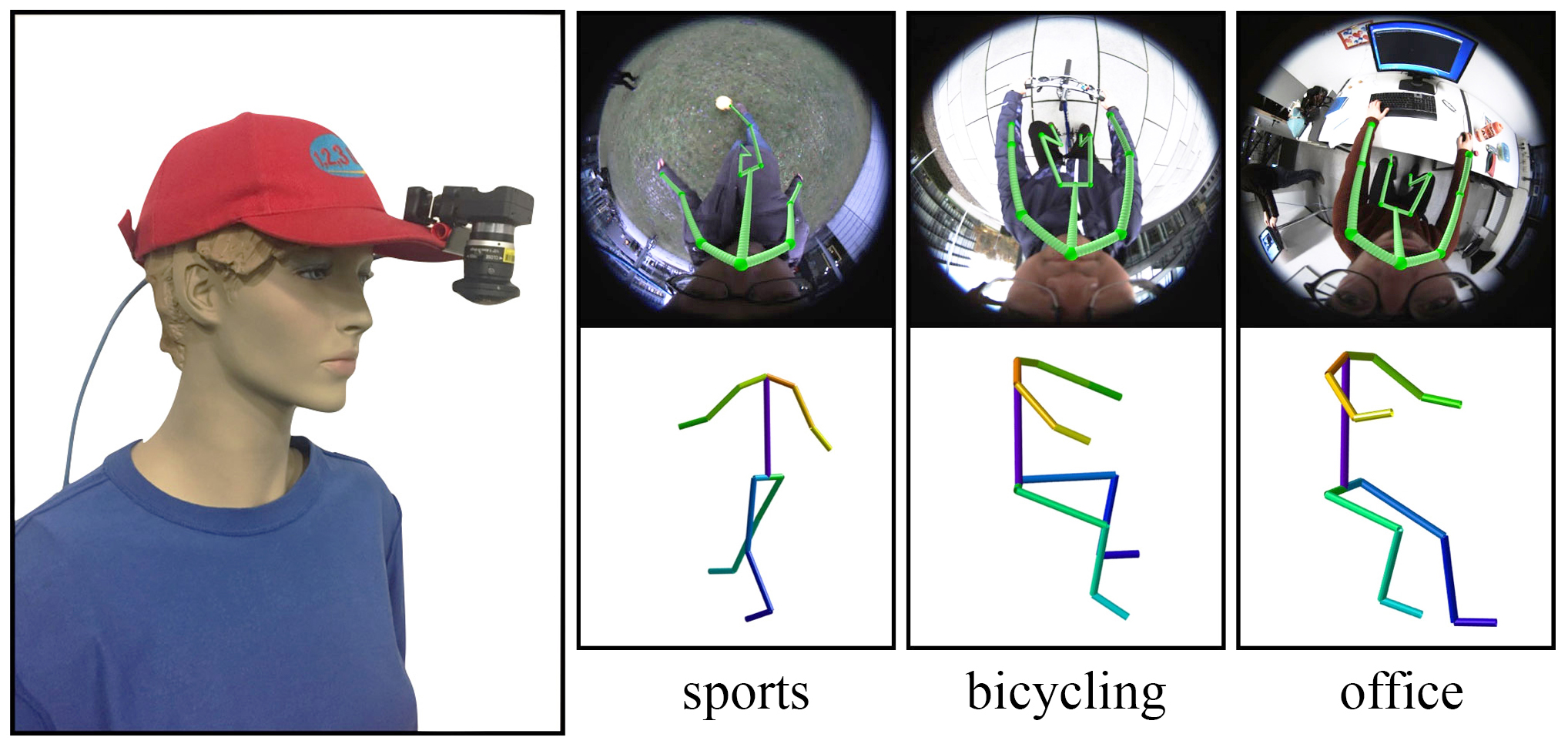}
  \caption{Our real-time mobile 3D pose estimation system is based on a single monocular cap-mounted fisheye camera that is attached to a standard baseball cap.
  	The setup is lightweight and enables 3D pose estimation in everyday situations.}
	\label{fig:teaser}
}

\vgtcinsertpkg

\begin{document}
	
\maketitle

\section{Introduction}
\label{sec:intro}

The goal of this work is to solve the problem of \emph{mobile} 3D human pose estimation in a wide range of activities performed in unconstrained real world scenes, such as walking, biking, cooking, doing sports and office work.
The resulting 3D pose can be used for action recognition, motion control, and performance analysis in fields such as sports, animation and health-care.
A real-time solution to this problem is also desirable for many virtual reality (VR) and augmented reality (AR) applications.

Such 3D human pose estimation in daily real world situations imposes a unique set of requirements on the employed capture setup and algorithm, such as: mobility, real-time performance, robustness to long capture sequences and fast recovery from tracking failures.
In the past, many works for outside-in 3D human pose estimation have been proposed, which use a single or multiple cameras placed statically around the user~\cite{mono-3dhp2017,VNect_SIGGRAPH2017,ElhayAJTPABST2015,RhodiRRST2015,Pavlakos17,Simon17}.
However, daily real world situations make outside-in capture setups impractical, since they are immobile, can not be placed everywhere, require a recording space without occluders in front of the subject, and have only a small recording volume.

Motion capture systems based on body-worn sensors, such as inertial measurement units (IMUs) \cite{von2017sparse} or multi-camera structure-from-motion (SFM) from multiple limb-mounted cameras \cite{ShiraPSSH2011}, support mobile capturing. 
However, these setups are expensive, require tedious pre-calibration, and often require pose optimization over the entire sequence, which prevents real-time performance.
Most closely related to our approach is the EgoCap~\cite{Rhodin:2016} system that is based on two head mounted fisheye cameras.
While it alleviates the problem of a limited capture volume, the setup is quite heavy and requires uncomfortable, obtrusive large extension-sticks.
EgoCap also requires dedicated 3D actor model initialization based on keyframes, does not run at real-time rates for the full body, and has not been shown to be robust on very long sequences.

In contrast, we tackle the unique challenges of real-time ubiquitous mobile 3D pose estimation with a novel lightweight hardware setup (see~Fig.~\ref{fig:teaser}) that converts a standard baseball cap to a device for accurate 3D human pose estimation using a single fisheye camera.
Our approach fulfills all requirements mentioned at the outset:
1) Our hardware setup is compact, lightweight and power efficient, which makes it suited for daily mobile use.
2) Our approach requires no actor calibration and works for general and dynamic backgrounds, which enables free roaming during daily activities.
3) From the live stream of the cap-mounted camera, our approach estimates 3D human pose at 60~Hz.
4) Our online frame-by-frame pose estimation solution is suitable for capturing long sequences and automatically recovers from occasional failures.

\begin{figure}[t]
	\begin{center}
		\includegraphics[width=0.8\linewidth]{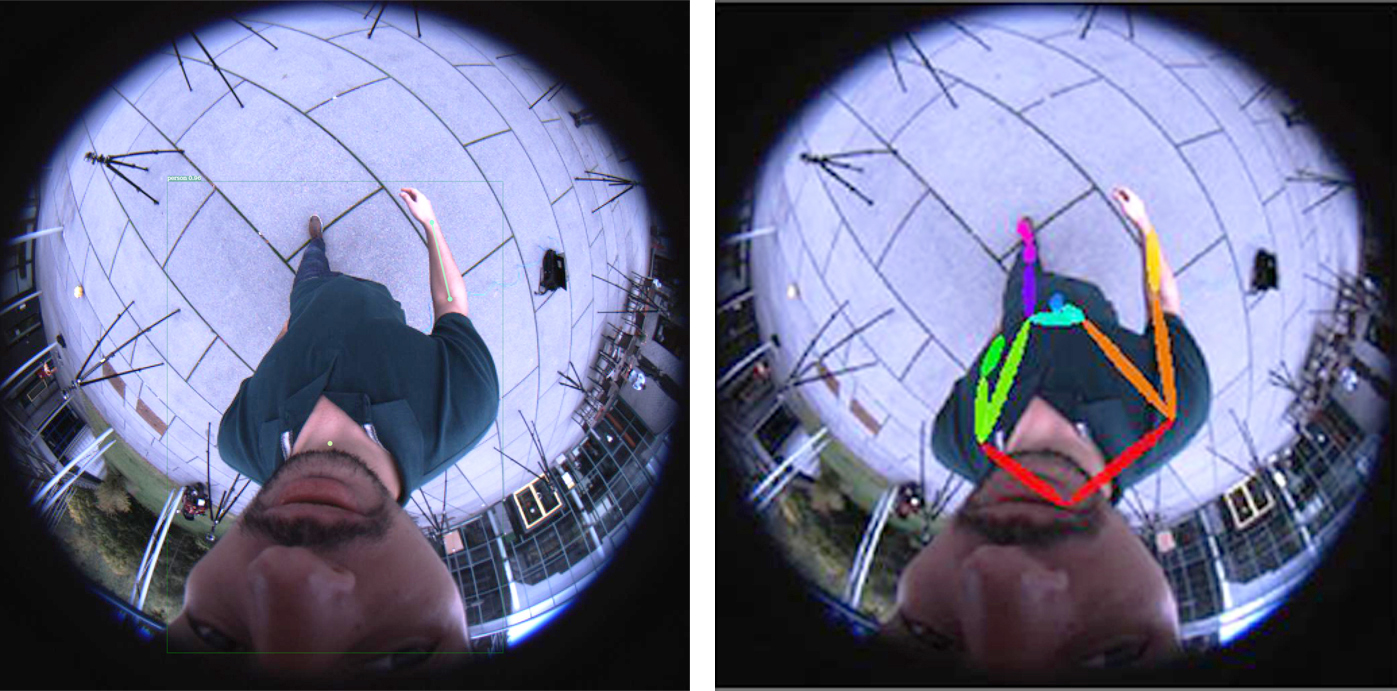}
	\end{center}
	\vspace{-0.3cm}
	\caption
	{
		The state-of-the-art 2D human pose estimator Mask R-CNN~\cite{He:2017:MaskRCNN} trained on the COCO dataset~\cite{ms-coco} fails on images captured by our setup (left). Our 2D pose estimation results (right).
	}  
	\vspace{-0.4cm}
	\label{fig:maskrcnn}
\end{figure}

As is true for most of the recent outside-in monocular 3D human pose estimation methods, our approach is also based on a deep neural network.
However, existing methods do not apply well to our setting.
First, their training data is captured with regular cameras and mostly from chest high viewpoints.
Thus, they fail on our images, which are captured from a top-down view and exhibit a large radial distortion (see Fig.~\ref{fig:maskrcnn}).
Second, most of the existing methods directly estimate 3D human pose in the form of 3D joint locations relative to the pelvis and do not respect the 2D-3D consistency.
This not only makes them yield bad 2D overlay of 3D pose results on the images, but also makes the 3D pose estimation less accurate, since even a small 2D displacement translates to a large 3D error due to the short focal length of the fisheye camera.
Third, the close proximity of the camera to the head creates a strong perspective distortion, resulting in a large upper body and very small lower body in the images, which makes the estimation of the lower body less accurate.
To solve these problems, we propose a new ground truth training corpus of top-down fisheye images and, more importantly, a novel 3D pose estimation algorithm based on a CNN that is specifically tailored to the uniqueness of our camera position and optics.
Specifically, instead of directly regressing the 3D joint locations, we disentangle the 3D pose estimation problem to the following three subproblems: 
1) 2D joint detection from images with large perspective and radial distortions, which is solved with a two-scale location invariant convolutional network,
2) absolute camera-to-joint distance estimation, which is solved with a location sensitive distance module that exploits the spatial dependencies induced by the radial distortion and fixed camera placement relative to the head and
3) recovering the actual joint position by back-projecting the 2D detections using the distance estimate and the optical properties of the fisheye lens.
Our disentangled approach leads to not only accurate 3D pose estimation, but also good 2D overlay of results, since, by construction, the 3D joint locations will exactly re-project to the corresponding 2D detections.

To the best of our knowledge, our work is the first approach that performs real-time mobile 3D human pose estimation from a single egocentric fisheye camera.
Our qualitative and quantitative evaluations demonstrate that the proposed approach outperforms the baseline methods on our test set.
Our datasets and code are publicly available at \url{http://gvv.mpi-inf.mpg.de/projects/wxu/Mo2Cap2/}.

\section{Related Work}
\label{sec:rel-work}

In the following, we categorize relevant motion capture approaches in terms of the employed setup.

\paragraph{Studio and Multi-view Motion Capture}
Multi-view motion capture in a studio typically employs ten or more cameras.
For marker-based systems the subject has to be instrumented, \eg,  with a marker or LED suit.
Marker-less motion-capture algorithms overcome this constraint \cite{BreglM1998,TheobASST2010,MoeslHKS2011,HolteTTM2012,urtasun2006temporal,Gall:2010,SigalBB2010,SigalIHB2012,StollHGST2011,JooLTGNMKNS2015}, with recent work \cite{AminARS2009,BurenSC2013,ElhayAJTPABST2015,RhodiRRST2015,Robertini:2016,Pavlakos17,Simon17} even succeeding in outdoor scenes and using fewer cameras. 
The static camera setup ensures high accuracy but imposes a constrained recording volume, has high setup time and cost, and breaks when the subject is occluded in crowded scenes.
On the other hand, mobile hand-held solutions require a team of operators \cite{hasler2009markerless,wang2017outdoor,ye2012performance}.
In contrast, our system does not require any additional operators than the user, a multi-camera setup or complicated synchronization and calibration of multi-camera systems.
This makes our system more practical in everyday situations.

\paragraph{Monocular Human Pose Estimation}
Monocular human pose estimation is a requirement for many consumer-level applications.
For instance, human-computer interaction in living-room environments was enabled by real-time pose reconstruction from a single RGB-D camera \cite{Shotton:2011,Baak:2011,Wei:2012}.
However, active IR-based cameras are unsuitable for outdoor capture in sunlight and their high energy consumption limits their mobile application.
Purely RGB-based monocular approaches for capture in more general scenes have been enabled with the advent of convolutional neural networks (CNNs) and large training datasets \cite{Ionescu14a,Rogez16,ChenWLSTLCC2016,varol17}.
Methods either operate directly on images \cite{Li14a,tekin_structured_bmvc16,Zhou16b,mono-3dhp2017,Tekin17a}, lift 2D pose detections to 3D \cite{Bogo16,Zhou16a,Chen2016,Yasin16,Jahangiri17},
or use motion compensation and optical flow in videos \cite{Tekin:2016,alldieck2017optical}.
The most recent improvements are due to hierarchical processing \cite{Tome17,Popa17}
and 
combining 2D and 3D tasks~\cite{pavlakos17volumetric,VNect_SIGGRAPH2017,zhou2017weakly}.
Our approach is inspired by the separation of 2D pose and depth estimation by \cite{zhou2017weakly}, which, however, assumes an orthographic projection model that does not apply to the strong distortion of our fisheye-lens and is different in that it predicts relative, hip-centered depth instead of absolute distance.
While these approaches enable many new applications, the camera is either fixed, which imposes a restricted capture volume, or needs to be operated by a cinematographer that follows the action.
We build upon these monocular approaches. We generalize them to a head-mounted fisheye setup and address its unique challenges, such as the special top-down view and the large distortion in the images.
The robustness and accuracy is significantly improved compared to the state-of-the-art
by a new training dataset and by exploiting the characteristics of the head-mounted camera setup with a disentangled 3D pose estimation approach.

\paragraph{Body-worn Motion Sensors}
For some studies, the restricted capture volume of static camera systems is overcome by using inertial measurement units (IMUs) \cite{TautgZKBWHMSE2011,von2017sparse} or exoskeleton suits (\eg, METAmotion Gypsy).
These form an inside-in arrangement, where the sensors are body-worn and capture body motion independent of external devices.
Unfortunately, the sensor instrumentation and calibration of the subject cause long setup times and makes capturing multitudes of people difficult. Furthermore, IMU measurements require temporal integration to obtain position estimates, which is commonly addressed by offline batch-optimization to minimize drift globally \cite{von2017sparse}. We aim at lower setup times and real-time reconstruction with minimal latency, \eg, for interactive virtual reality experiences.

\paragraph{Mobile Motion Capture}
Self-contained motion capture in every-day conditions demands for novel concepts.
By attaching 16 cameras to the subject's limbs and torso in an inside-out configuration Shiratori \etal recover the human pose by structure from motion on the environment, enabling free roaming in static backgrounds \cite{ShiraPSSH2011}.
For dynamic scenes, vision-based inside-in arrangements have been proposed.
The camera placement is task specific.
Facial expression and eye gaze have been captured with a helmet-mounted camera or rig \cite{JonesFYMBIBD2011,WangCF2016,Sugano:2015}, hand articulation and action from head-mounted \cite{SridhMOT2015,singh2017trajectory,wu2017yolse} or even wrist- or chest-worn cameras \cite{Kim:2012,Rogez2014}.
The user's gestures and activity can also be recognized from a first-person perspective \cite{FathiFR2011,kitani2011fast,OhnisKKH2016,ma2016going,cao2017egocentric}.

However, capturing accurate full body motion in such a body-mounted inside-in camera arrangement is considerably more challenging, as it is difficult to observe the whole body from such close proximity.
Yonemoto \etal propose indirect inference of arm and torso poses from arm-only RGB-D footage \cite{YonemMOSST2015} and Jiang attempted
to reconstruct full-body pose by analyzing the egomotion and observed scene \cite{JiangG2016}, but indirect predictions have low confidence and accuracy.
A first approach towards direct full-body motion capture from the egocentric perspective was proposed by Rhodin \etal~\cite{Rhodin:2016}.
A 3D kinematic skeleton model is optimized to explain 2D features in each of the views of a stereo fisheye camera mounted on head-extensions similar in structure to a selfie stick.
While enabling free roaming many application scenarios are hampered by the bulky stereo hardware.
This approach achieves interactive framerates for upper body tracking only, while ours enables real-time 3D pose estimation for the full human body.
To be less intrusive, we propose a lightweight hardware setup based on a single cap-mounted fisheye camera, which requires an entirely different reconstruction algorithm since the optimization used by Rhodin \etal intrinsically requires stereo vision with a large baseline.
Furthermore, our algorithm is capable of estimating 3D pose from a single frame, which reduces the chance of long-term tracking failures and enables capture of arbitrarily long sequences without manual intervention.

\section{The $\textbf{Mo}^\textbf{2}\textbf{Cap}^\textbf{2}$ Approach}
\label{sec:method}

$\textnormal{Mo}^\textnormal{2}\textnormal{Cap}^\textnormal{2}$ is a real-time approach for mobile 3D human body pose estimation based on a single cap-mounted fisheye camera.
Our headgear augments a standard baseball cap with an attached fisheye camera.
It is lightweight, comfortable and very easy to put on.
However, the usage of only one camera view, the very slanted and proximate viewpoint and the fisheye distortion makes 3D pose estimation extremely challenging.
We address these challenges by a novel disentangled 3D pose estimation algorithm based on a CNN that is specifically tailored to our setup.
We also contribute a large scale training corpus of synthetic top-down view fisheye images with ground truth annotations.
It covers a wide range of body motion and appearance.
In the following, we provide more details on these aspects.

\subsection{Lightweight Hardware Setup}

Our work is the first approach that performs 3D real-time human body pose estimation from a single head-mounted camera.
Previous work \cite{Rhodin:2016} has demonstrated successful motion capture with a helmet-mounted fisheye stereo pair.
While their results are promising, their setup has a number of practical disadvantages.
Since they mount each of the cameras approximately 25 cm away from the forehead, the weight of the two cameras translates into a large moment, making their helmet quite uncomfortable to wear.
Furthermore, their large stereo baseline of 30-40 cm in combination with the large forehead-to-camera distance forces the actor to stay far away from walls and other objects, which limits usability of the approach in many everyday situations.  

In contrast, our setup is based on a single fisheye camera mounted to the brim of a standard baseball cap (see Fig.~\ref{fig:teaser}), which leads to a lightweight, comfortable and easy-to-use head-gear.
Installed only 8cm away from the head, the weight of our camera (only 175g) translates to a very small moment, which makes our setup practical for many scenarios.
Note, there exist even smaller/lighter cameras we could use, without making any algorithmic changes to our method.
One could even integrate the small camera inside the brim, which would make the setup even lighter.
Such engineering improvements are possible, but beyond the scope of this paper. 
Our fisheye camera has a $182^\circ$ field of view in both the horizontal and vertical direction.
This allows capturing the full body under a wide range of motion, including fully extended arms.
However, our hardware setup also makes 3D pose estimation more challenging since 1) explicit depth is not available in our monocular setup and 2) due to the shorter forehead to camera distance, the body is viewed quite obliquely.
Solving 3D pose estimation under such challenging conditions is the key contribution of our paper.

\subsection{Synthetic Training Corpus}

We now present our egocentric fisheye training corpus that enables training of a deep neural network that is tailored to our unique hardware setup.
Capturing a large amount of annotated 3D pose data is already a mammoth task for outside-in setups and it is even harder for egocentric data.
Since manual labeling in 3D space is impractical, \cite{Rhodin:2016} proposes to use marker-less multi-view motion capture with externally mounted cameras to get 3D annotations.

However, even with the help of such professional motion capture systems, acquiring a large number of annotated real-life training examples for the egocentric viewpoint is still a time consuming and tedious recording task.
It requires to capture the training data in a complex multi-view studio environment and precise 6 DOF tracking of the cap-mounted camera, such that the 3D body pose can be reprojected to the egocentric viewpoint of interest.
Furthermore, scalability to general scenes requires foreground/background augmentation, which typically relies on extra effort of capturing with green screen and image segmentation with color keying.
Given the difficulty of capturing a large amount of training data, the EgoCap \cite{Rhodin:2016} system does not scale to a large corpus of motions and real world diversity of human bodies in terms of shape and appearance, as well as diversity of real scene backgrounds.
Furthermore, their dataset cannot be directly used for our method, due to the different camera position relative to the head.

\begin{figure}[t]
	\begin{center}
		\includegraphics[width=0.85\linewidth]{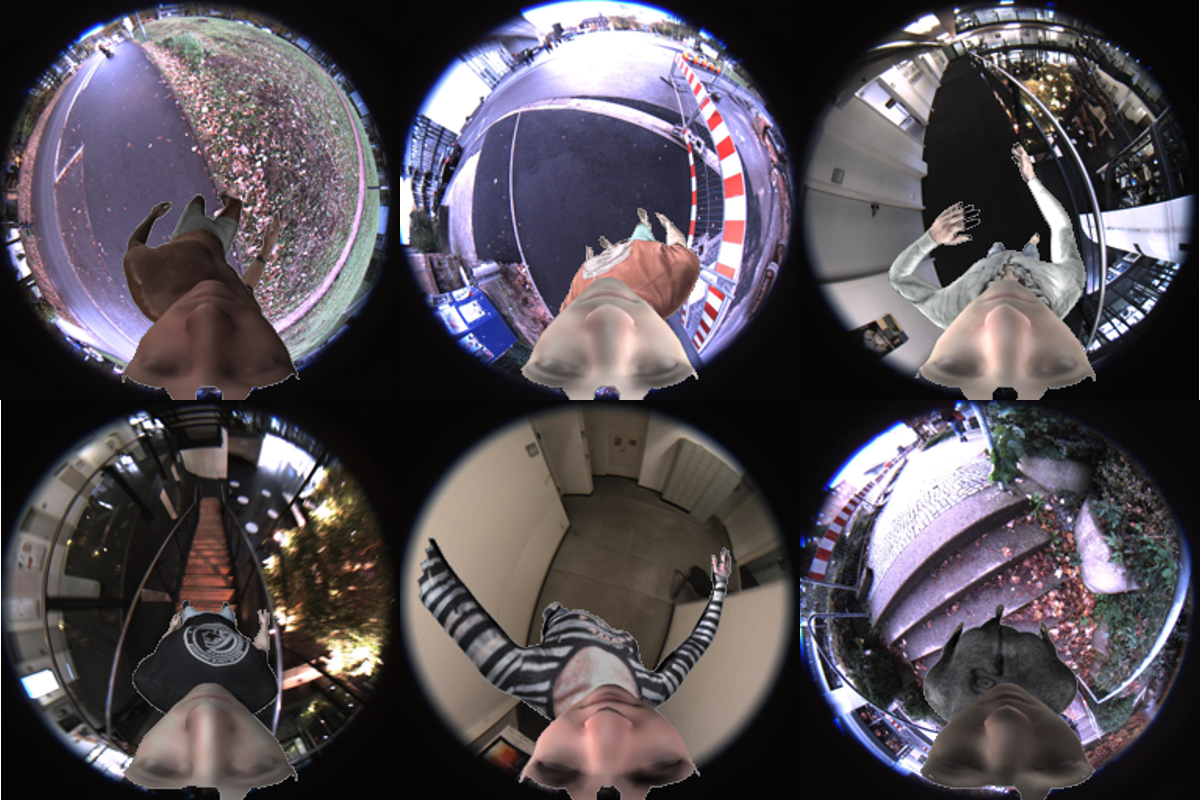}
	\end{center}
	\vspace{-0.4cm}
	\caption
	{
		Example images of our synthetically rendered fisheye training corpus.
		Our synthetic training corpus features a large variety of poses, human body appearance and realistic backgrounds.
	}  
	\vspace{-0.4cm}
	\label{fig:dataset}
\end{figure}

\begin{figure*}[t]
	\begin{center}
		\includegraphics[width=0.95\linewidth]{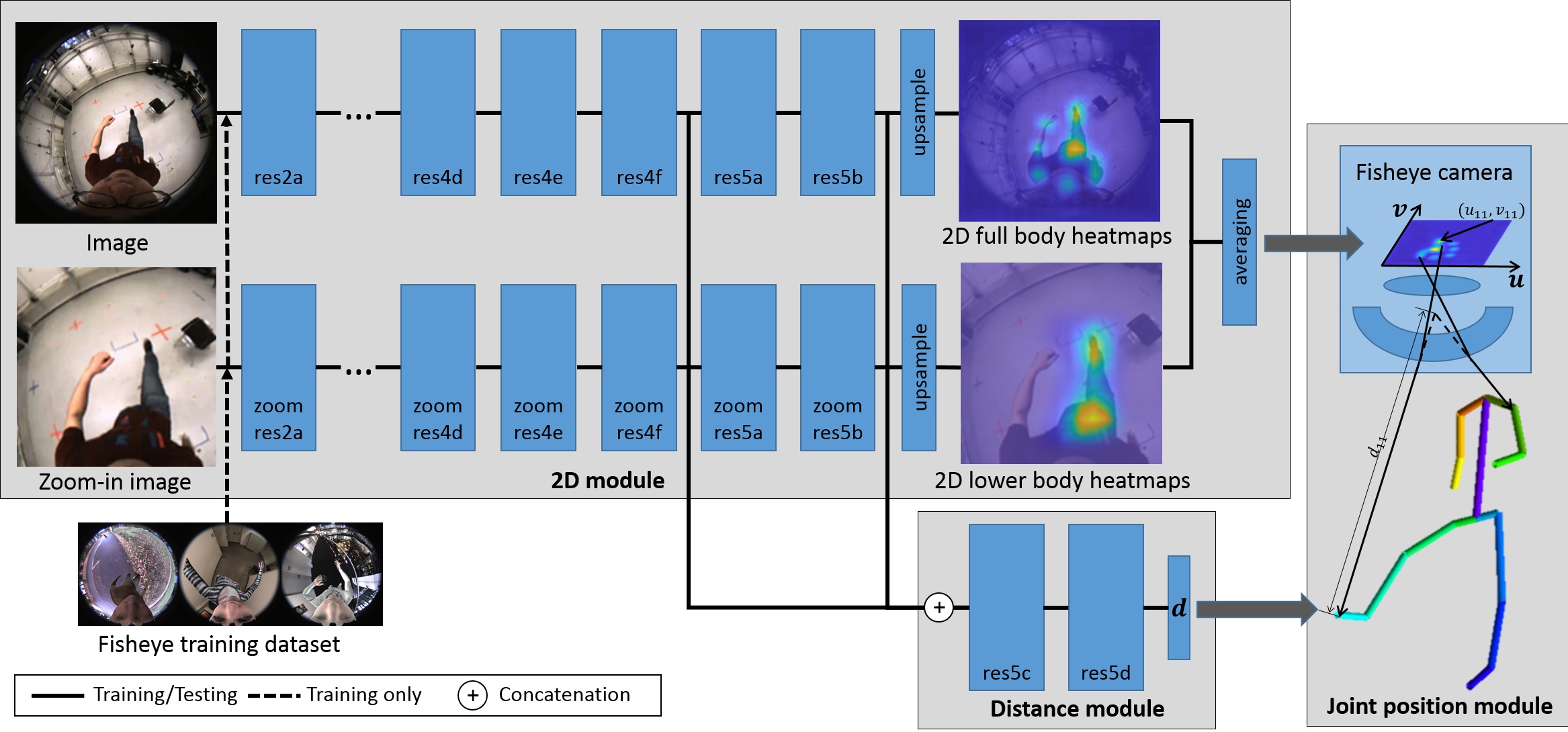}
	\end{center}
	\vspace{-0.5cm}
	\caption{
		Our disentangled 3D pose estimation method, which is specifically tailored to our cap-mounted fisheye camera setup, consists of three modules: 
		The two branched \textbf{2D module} estimates the 2D joint location heatmaps of the full body from the original image and the lower body from the zoom-in image.
		The \textbf{distance module} estimates the distance between the camera and each joint.
		The \textbf{joint position module} recovers the actual joint position by back-projecting the 2D detection using the joint-to-camera distance estimate and the intrinsic calibration of the fisheye camera.
	}	
	\vspace{-0.3cm}
	\label{fig:CNN}
\end{figure*}

In contrast, we alleviate these difficulties by rendering a synthetic human body model from the egocentric fisheye view.
Note that the success of any learning based method largely depends on how well the training corpus resembles the real world in terms of motion, body appearance and environment realism.
Therefore, care must be taken to ensure that 1) the variety of motion and appearance is maximized and 2) that the differences between synthetic and real images are minimized.
On one hand, to achieve a large variety of training examples, we build our dataset on top of the large scale synthetic human SURREAL dataset~\cite{varol17}.
We animate characters using the SMPL body model \cite{SMPL:2015} with uniformly sampled motions from the CMU MoCap dataset \cite{cmu_mocap}.
Body textures are chosen randomly from the texture set provided by the SURREAL dataset \cite{varol17}.
In total, we render 530,000 images (see Fig.~\ref{fig:dataset}), which encompass around $3000$ different actions and more than $700$ different body textures.
On the other hand, to generate realistic training images, we mimic the camera, lighting and background of the real world scenario.
Specifically, images are rendered from a virtual fisheye camera attached to the forehead of the character at a distance similar to the size of the brim of the used real world baseball cap.
To this end, we calibrate the real world fisheye camera using the omni-directional camera calibration toolbox \textit{ocamcalib} \cite{scaramuzza2006toolbox} and apply the intrinsic calibration parameters to the virtual camera. 
Characters are rendered using a custom shader that models the radial distortion of the fisheye camera.
Note that the camera position with respect to the head might change slightly, due to the camera movements and varying wearing angles and positions of the cap.
To simulate this effect, we add a random perturbation to the virtual fisheye camera position.
Random spherical harmonics illumination is used with a special parameterization to ensure a realistic top down illumination.
All images are augmented with the backgrounds chosen randomly from a set of more than 5000 indoor and outdoor ground plane images captured by our fisheye camera.
To gather such background images, we attach the fisheye camera to a long stick to obtain images that do not show the person holding the camera.
Furthermore, we applied a random gamma correction to the rendered images, such that the network becomes insensitive to the specific photometric response characteristics of the used camera.

Our synthetic dataset contains the ground truth annotation of 2D and 3D joint positions, which can be easily generated using the body model and the camera calibration.
Specifically, we provide the joint positions of the following 15 body joints: \textit{neck, shoulders, elbows, wrists, hips, knees, ankles and toes}.
The 3D joint positions are with respect to the fisheye camera coordinate system for our egocentric setup.
The joint-to-camera distances are computed based on the 3D joint position (see Sec.~\ref{sec:pose_est}).
The 2D joint position annotation is provided in the form of 2D heatmaps.
To this end, we first project the ground truth 3D joint positions onto the image space using the camera calibration.
The projections are then resized to a resolution of $64\times 64$, and we put a Gaussian kernel of size $5\times 5$ and standard deviation $0.8$ at each of the 2D joint positions. 
Finally, the 2D heatmaps are further downsampled to $32\times 32$.

\subsection{Monocular Fisheye 3D Pose Estimation}
\label{sec:pose_est}
Our disentangled 3D pose estimation method consists of three modules (see Fig.~\ref{fig:CNN}).

The \textbf{2D module} of our method estimates 2D heatmaps of the joint locations in image space, where we adopt a fully convolutional architecture that is suited for 2D detection problems.
As mentioned before, the strong perspective distortion of our setup makes the lower body appear particularly small in the images and therefore leads to lower accuracy in the estimation of the lower body joints.
To solve this problem, we propose a 2D pose estimation module consisting of two independently trained branches, which see different parts of the images.
The original scale branch sees the complete images and predicts the 2D pose heatmaps of 15 joints in the full body.
The zoom-in branch only sees the $2\times$ zoomed central part of the original images.
This zoom-in branch predicts the 2D heatmaps of the 8 lower body joints (hips, knees, ankle and toes), since these joints project into this central region in most of the images captured by our cap-mounted camera.
Our zoom-in branch yields more accurate results on the lower body than the original scale branch, since it sees the images at $2\times$ higher resolution.
The lower body heatmaps from the two branches are then averaged.

The \textbf{distance module} performs a vectorized regression of per-joint absolute camera space depth, \ie, the distance between the camera and each joint, based on the higher and medium level features of the 2D module.
In contrast to the fully convolutional architecture of our 2D module, here we use a fully connected layer that can exploit the spatial dependencies in our setup induced by the radial distortion and fixed camera placement relative to the head.
Please note that absolute distance estimation is not practical for the classical outside-in camera setup, where the subject is first cropped in 2D to a normalized pixel scale from which 3D pose is estimated, by which absolute scale information is lost.

\begin{figure*}[ht!]
	\begin{center}
		\includegraphics[width=0.85\linewidth]{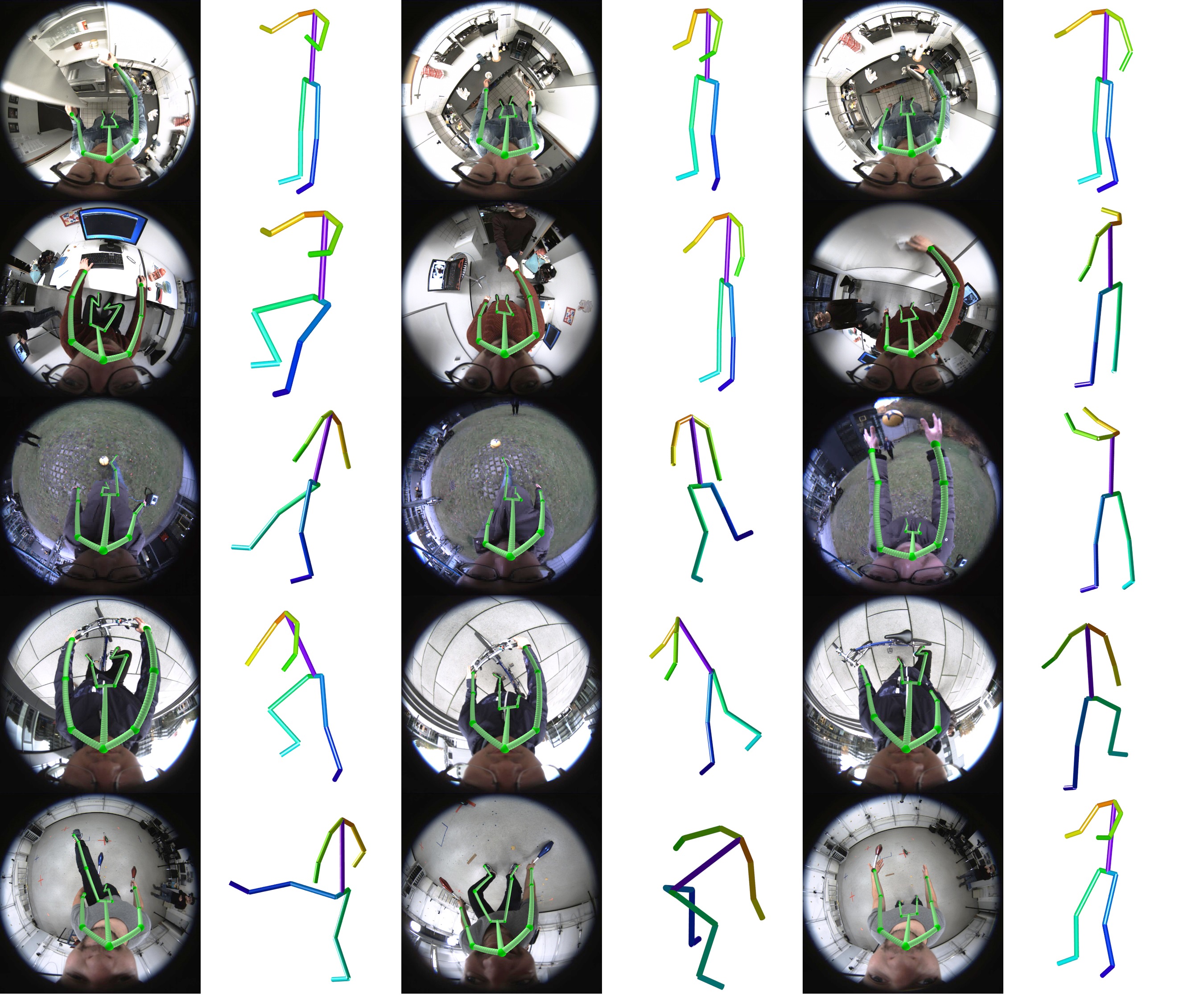}
	\end{center}
	\vspace{-0.3cm}
	\caption
	{
		Results in a variety of everyday situations. Left: our 3D pose results overlaid on the input images; Right: our 3D pose results from a side view.
	}  
	\vspace{-0.3cm}
	\label{fig:qualitative_live}
\end{figure*}

At last, the \textbf{joint position module} recovers the actual joint position by back-projecting the 2D detections using the distance estimate and the intrinsic calibration (including the distortion-coefficients) of the fisheye camera.
To this end, we first read out the $u,v$ coordinates of each joint from the averaged heatmaps.
Then, given the calibration of the fisheye camera \cite{scaramuzza2006toolbox}, each 2D joint detection $[u,v]^T$ can be mapped to its corresponding 3D ray vector $[x,y,z]^T$ with respect to the fisheye camera coordinate system:
\begin{equation}
\left[\begin{array}{c}x\\y\\z\end{array}\right] = \left[\begin{array}{c}u\\v\\f(\rho)\end{array}\right]
\; ,
\end{equation}

where $\rho=\sqrt{u^2+v^2}$, $f(\rho) = \alpha_0+\alpha_1\rho+\alpha_2\rho^2+\alpha_3\rho^3+\alpha_4\rho^4+\dots$ is a polynomial function that is obtained from camera calibration.
The 3D position of each joint $P$ is obtained by multiplying the direction vector with the predicted absolute joint-to-camera distance $d$,
\begin{equation}
P = \frac{d}{\sqrt{x^2+y^2+z^2}}\left[x,y,z\right]^T.
\end{equation}

Our disentangled 3D pose estimation method ensures that the 3D joint location will exactly reproject to its 2D detection, handles the scale difference between upper and lower body and leverages location dependent information of the egocentric setup as a valuable depth cue, and therefore results in more accurate 3D pose estimation than previous architectures trained on the same data.

\paragraph{Implementation of our network}
Each branch of our 2D module consists of $15$ residual blocks \cite{resnet_He2015} and performs a deconvolution and two convolutions to upsample the prediction to the heatmap size of $32\times 32$ pixels given images of resolution $256\times 256$ pixels as input.
In addition to the euclidean $\ell_2$ loss on the final heatmap predictions, we add two additional intermediate supervision losses (after $11$ and $14$ residual blocks) for faster convergence during training and to prevent vanishing gradients during back-propagation.
The architecture of the distance module is based on $2$ additional residual blocks, $2$ convolution and $1$ fully connected layer.
We concatenate the output features of the 13th and 15th residual blocks of the two 2D module branches, and pass it to the distance module.

\paragraph{Multi-stage Training}

Our training corpus is based on synthetically rendered images.
To make our network better generalize to real world imagery, we train it in multiple stages using transfer learning.
First, we pre-train the 2D module of our network on an outside-in pose estimation task based on the MPII Human Pose~\cite{AndriPGS2014} and LSP~\cite{JohnsonCVPR2011} datasets.
These real images with normal optics enable our network to learn good low-level features, which, at that feature level, are transferable to our egocentric fisheye setup.
Afterwards, we fine tune the two branches of the 2D module separately on the images from our synthetically rendered fisheye training corpus and the $2\times$ zoomed version of them respectively.
Note that in order to preserve the low level features learned from real images, we decrease the learning rate multiplier to $0.001$ for the initial $13$ residual blocks.
Afterwards, we fix the weights of the 2D module and train our distance module.
The Euclidean loss is used for the final loss and all intermediate losses.
In all training stages, we use a batch size of $24$, and we train the 2D module for $50$k iterations, and the distance module for $70$k iterations.
For the fine tuning stages, we use a learning rate of 0.05.
AdaDelta is used for optimization~\cite{adadelta}.

\begin{figure*}[!ht]
	\begin{center}
		\includegraphics[width=0.85\linewidth]{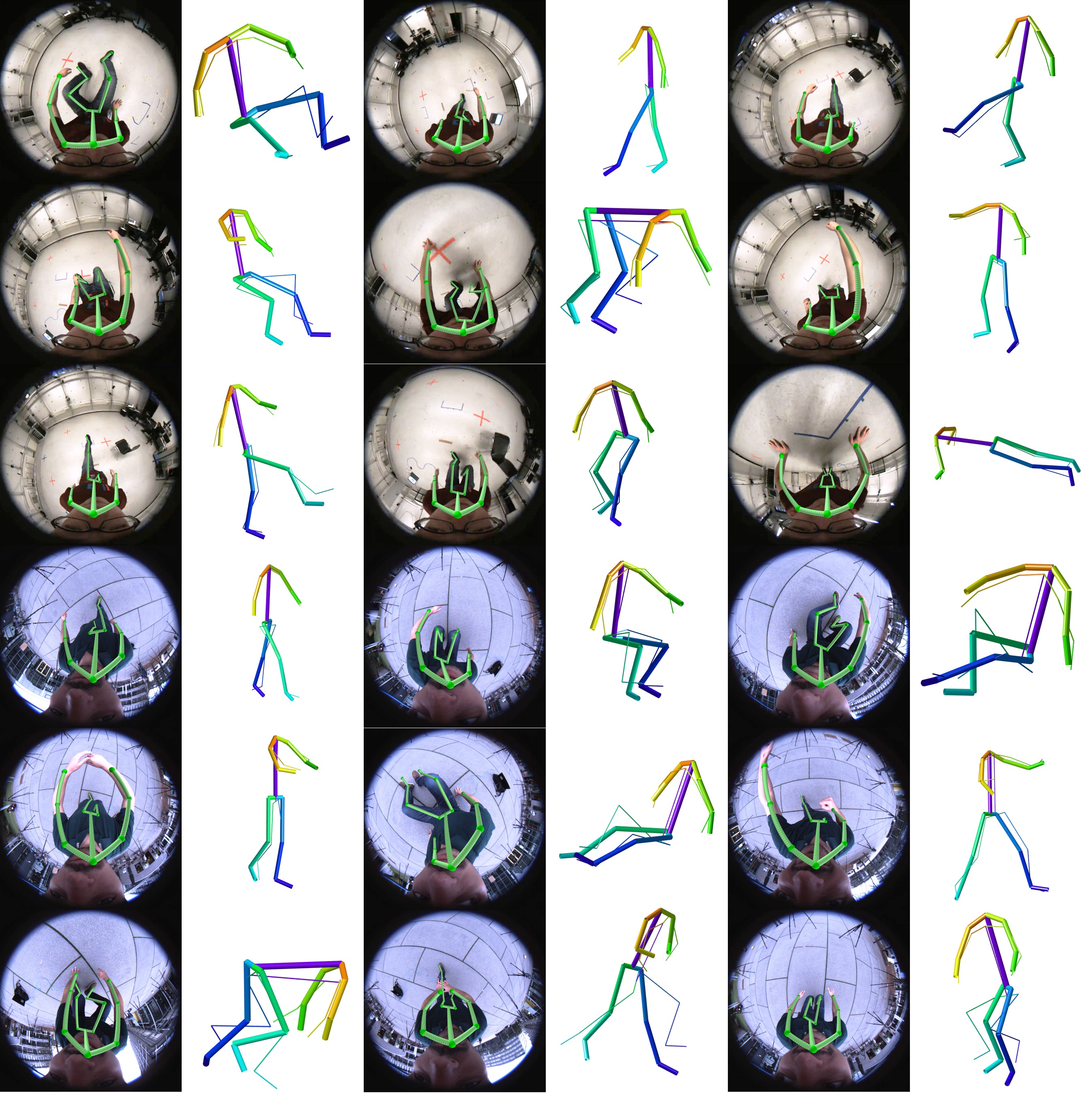}
	\end{center}
     \vspace{-0.3cm}
	\caption
	{
		Results on the indoor and outdoor sequences with ground truth. Left: 3D pose results overlaid on the input images; Right: 3D pose results from a side view, the thinner skeleton is the ground truth obtained using a commercial multi-view motion capture software.
	}  
	%		\vspace{-0.5cm}
	\label{fig:qualitative}
\end{figure*}

\section{Results}
\label{sec:results}

We study the effectiveness and accuracy of our approach in different scenarios.
Our $\textnormal{Mo}^\textnormal{2}\textnormal{Cap}^\textnormal{2}$ system runs at 60~Hz on an Nvidia GTX 1080 Ti, which boils down to 16.7~ms for the forward pass.
Thus, our approach can be applied in  many applications in which real-time performance is critical, \eg,~for motion control in virtual reality.

In the following, we first evaluate our approach qualitatively and quantitatively.
Then, we demonstrate that our disentangled 3D human pose estimation approach leads to significant gains in reconstruction accuracy.

\subsection{Qualitative Results}

Our lightweight and non-intrusive hardware setup allows the users to capture general daily activities.
To demonstrate this, we captured a test set of 5 activities, including both everyday and challenging motions, in unconstrained environments:
1) making tea in the kitchen,
2) working in the office,
3) playing football,
4) bicycling,
and 5) juggling.
Each sequence contains approximately 2000 frames.
The sequences cover a large variety of motions and subject appearances (see Fig.~\ref{fig:qualitative_live} for examples).
We can see that our method estimates accurate 3D poses for all sequences.
Even interactions with other people or objects are captured, which is a challenge even for multi-view outside-in methods.
Note that we capture the bicycling and juggling sequences to provide a comparison with the state-of-the-art egocentric 3D pose estimation approach of \cite{Rhodin:2016}, since they also show results for these two actions.
We can see that our monocular method yields comparable and sometimes more stable results than their binocular method.
Also note, in contrast to~\cite{Rhodin:2016}, our method runs in real-time on the full body, and does not require 3D model calibration of the user or any optimization as post-process.
The complete results on all sequences are shown in the supplementary video.

\subsection{Quantitative Results}

\begin{table*}[t]
\scriptsize
	\caption
	{
		Ground truth comparison on real world sequences.
		Our disentangled 3D pose estimation approach outperforms the vectorized 3D body pose prediction network of~\cite{mono-3dhp2017} and the location map approach used in~\cite{VNect_SIGGRAPH2017}, which are \textbf{trained on our dataset},  in terms of mean joint error (in mm) .
	}  
	%\vspace{-0.1cm}
	\centering
	\resizebox{1\textwidth}{!}{
		\begin{tabular}{ | l | c | c | c | c | c | c | c | c || c |}
			\hline
			\textbf{Indoor} & \textbf{walking} & \textbf{sitting} & \textbf{crawling} & \textbf{crouching} & \textbf{boxing} & \textbf{dancing} & \textbf{stretching} & \textbf{waving} & \textbf{total}\\ \hline \hline
			\textbf{3DV'17~\cite{mono-3dhp2017}}   & 48.7571  & 101.2177& 118.9554& 94.9254 & {57.3380} & {60.9604} & 111.3591& {64.4975} & 76.2813 \\ \hline
			\textbf{VNect~\cite{VNect_SIGGRAPH2017}} & 65.2818  & 129.5852& 133.0847& 120.3911& 78.4339& 82.4563& 153.1731& 83.9061& 97.8454 \\ \hline
			\textbf{Ours w/o zoom}  & {47.0895}  & {82.6745} & {98.9962} & {87.9168} & 58.7640& 63.6811& {109.2848} & 69.3515& {70.1923} \\ \hline
			\textbf{Ours  w/o averaging}  &45.8356	 & 77.6024	 & 99.9472	 & 83.8608	 & 55.2959	 & 60.5191	 & 115.7854	 & 66.972 & 	68.1455
			\\ \hline
			\textbf{Ours} &\textbf{38.4083}	&\textbf{70.9365}	&\textbf{94.3191}	&\textbf{81.898}	&\textbf{48.5518}	&\textbf{55.1928}	&\textbf{99.3448}	&\textbf{60.9205}	&\textbf{61.3977}
			\\ \hline
			\hline
			\textbf{Outdoor} & \textbf{walking} & \textbf{sitting} & \textbf{crawling} & \textbf{crouching} & \textbf{boxing} & \textbf{dancing} & \textbf{stretching} & \textbf{waving} & \textbf{total}\\ \hline \hline
			\textbf{3DV'17~\cite{mono-3dhp2017}}   & {68.6660}  &114.8663 &113.2263 &118.5457 &{95.2946} &{72.9855} &144.4816 &72.4117 &92.4635 \\ \hline
			\textbf{VNect~\cite{VNect_SIGGRAPH2017}} & 84.4322  &167.8719 &138.3871 &154.5411 &108.3584 &85.0144 &160.5673 & 96.2204 &113.7492 \\ \hline
			\textbf{Ours  w/o zoom}  & 69.3500  &{89.1967} &{99.7597} &{101.7018} &105.7102 &74.1185 &{134.5125} &{71.2431} &{87.3114} \\ \hline
			\textbf{Ours  w/o averaging} & 67.889 &	88.7139 & 99.2919 & 99.3326 & 106.3386 & 72.3075 & 136.4019 & 69.0395 & 86.3061 \\ \hline
			\textbf{Ours}  & \textbf{63.1027} &	\textbf{85.4761} & \textbf{96.6318} &	\textbf{92.8823} & \textbf{96.0142} &	\textbf{68.3541} &	\textbf{123.5616} &	\textbf{61.4151} &	\textbf{80.6366} \\ \hline 
		\end{tabular}
	}
	%	\vspace{-0.4cm}
	\label{tab:quantitative}
\end{table*}

Existing, widely used data sets for monocular 3D pose estimation, \eg, Human3.6M~\cite{Ionescu14a}, are designed for outside-in camera perspectives with normal optics, not our egocentric, body-worn fisheye setup. 
In turn, our absolute distance estimation without image cropping only applies to body-mounted scenarios.
In order to evaluate our method quantitatively, we therefore captured an extra test set with ground truth annotation containing 8 different actions across 5591 frames, recorded both indoors and outdoors with people in general clothing.
The recorded actions include walking, sitting, crawling, crouching, boxing, dancing, stretching and waving.
The 3D ground truth is recorded with a commercial external multi-view marker-less motion capture system~\cite{captury}.
Fig.~\ref{fig:qualitative} shows our 3D pose results overlaid on the input images (left) and from a side view (right), where the ground truth 3D pose is shown with the thinner skeleton.
Since our method does not estimate the global translation and rotation of the body, in order to quantitatively compare our method to the ground truth, we apply Procrustes analysis to register our results to the ground truth.
Following many other 3D pose estimation methods~\cite{VNect_SIGGRAPH2017,mono-3dhp2017}, we rescale the bone length of our estimated pose to the ``universal'' skeleton for quantitative evaluation.
The average per-joint 3D error (in millimeters) on different actions is shown in Tab.~\ref{tab:quantitative}.
Note that our accuracy is comparable with monocular outside-in 3D pose estimation approaches, even though our setting is much more challenging.

\subsection{Influence of the Network Architecture}

\begin{figure}[t]
	\begin{center}
		\includegraphics[width=\linewidth]{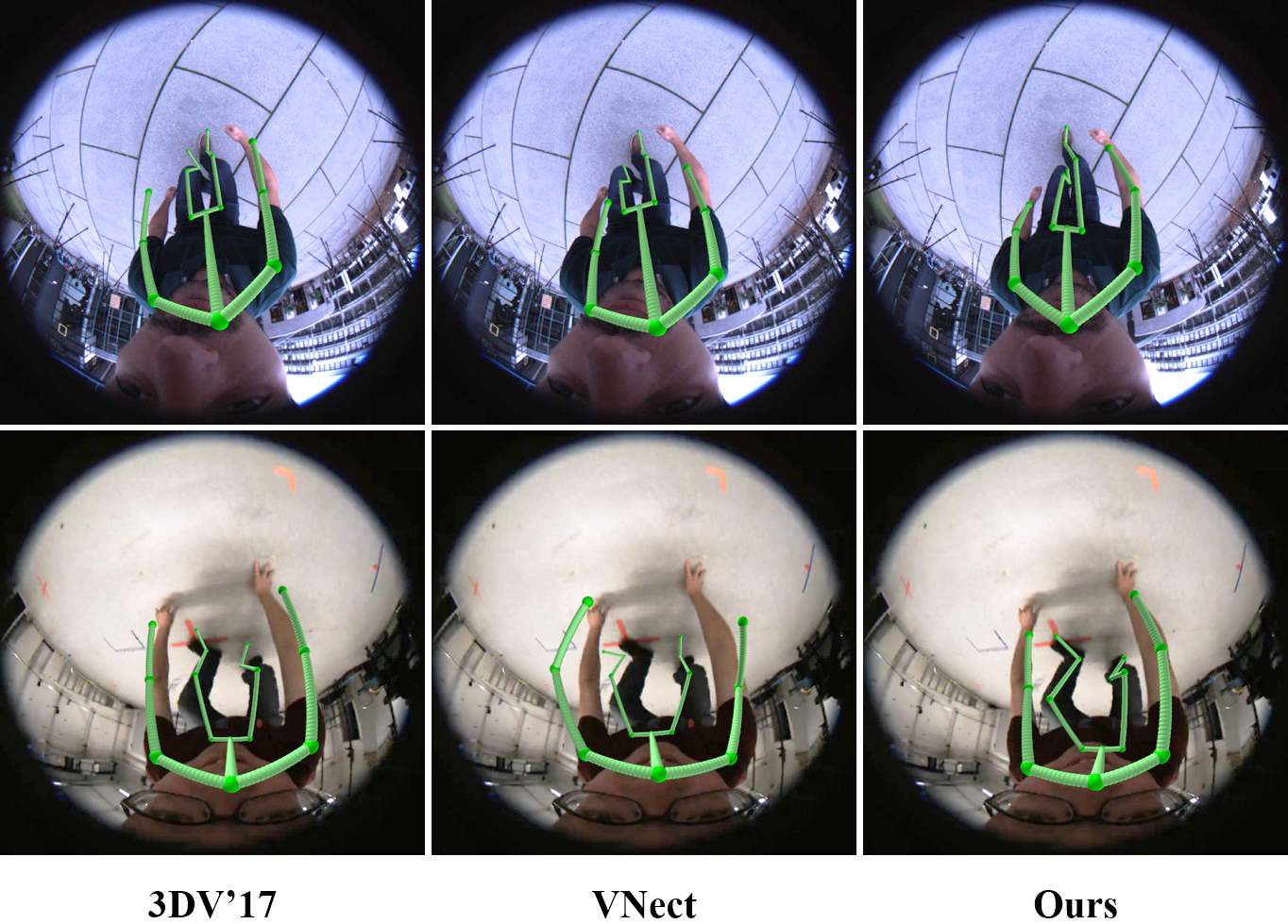}
	\end{center}
	\vspace{-0.3cm}
	\caption
	{
		Comparison of 3D pose results overlaid on the input images. Our results accurately overlay on the images, while the results of the baseline methods exhibit significant offsets.
	}  
	\label{fig:overlay}
	\vspace{-0.3cm}
\end{figure}

We also quantitatively compare our disentangled architecture to other state-of-the-art baseline approaches (see Tab.~\ref{tab:quantitative}) on our egocentric fisheye data. The latter were originally developed for outside-in capture from undistorted camera views.
Specifically, we compare to the vectorized 3D body pose prediction network of \cite{mono-3dhp2017} (referred to as \textbf{3DV'17}) and the location map approach used in \cite{VNect_SIGGRAPH2017} (referred to as \textbf{VNect}).
As all three methods are based on a ResNet, we modify their architectures to use the same number of ResNet blocks as ours for a fair comparison.
We also apply the same intermediate supervision to all three method and use the same training strategy.
We train all networks on our synthetic training corpus of egocentric fisheye images.
One can see that our disentangled 3D pose estimation approach outperforms these two state-of-the-art network architectures by a large margin (indoors: $19.5\%$, outdoors: $14.7\%$ over \textbf{3DV'17}) in terms of mean joint error (in mm), see~Tab.~\ref{tab:quantitative}.
This demonstrates that our architecture is especially well suited for our monocular fisheye setup.
In addition, our disentangled representation leads to good 2D overlay, since the 2D and 3D detections are consistent by construction.
A comparison of the 2D overlay results of the three different methods is shown in Fig.~\ref{fig:overlay}.
One can see that our 3D pose results accurately overlay on the images, while the results of the baseline methods exhibit significant offsets.

\begin{figure*}[t]
	\begin{center}
		\includegraphics[width=0.7\linewidth]{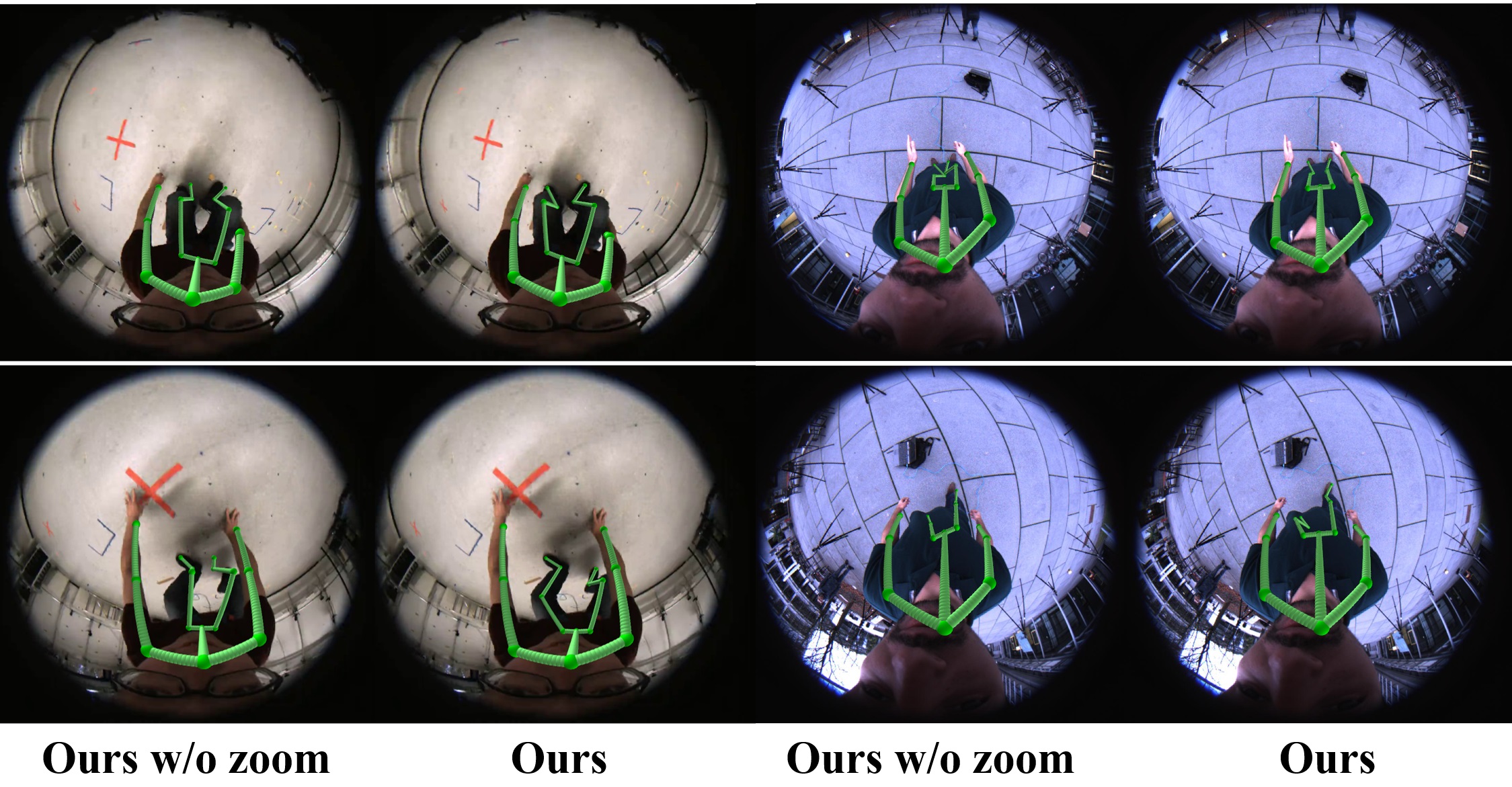}
	\end{center}
	\vspace{-0.6cm}
	\caption
	{
		Benefiting from the zoom-in branch, our full method yields significantly better overlay of the lower body joints.
	}  
	\label{fig:compare_zoom}
\end{figure*}

	\begin{figure}[t]
		\begin{center}
			\includegraphics[width=\linewidth]{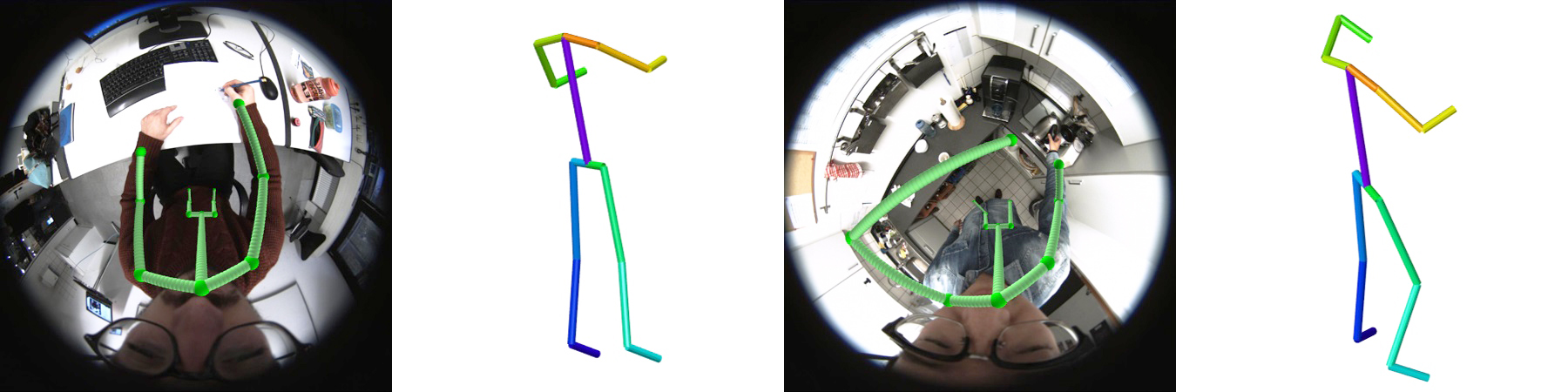}
		\end{center}
		\caption
		{
			Failure cases of our method. Left: Our method outputs a standing pose instead of a sitting pose, since the legs are completely occluded. Right: As the left arm is barely visible, our method aligns the arm to the edge of the cupboard.
		}  
		\label{fig:limitation}
		\vspace{-0.5cm}
	\end{figure}

We further perform an ablation study to evaluate the importance of the zoom-in branch of our 2D module.
We compare to two incomplete versions of our method: 1) with zoom-in branch completely removed (referred to as \textbf{Ours w/o zoom}) and 2) without averaging the heatmaps from the two branches, but only using those from the original scale branch (referred to as \textbf{Ours w/o averaging}).
We can see from Fig.~\ref{fig:compare_zoom} that, benefiting from the zoom-in branch,  our full method yields significantly better overlay of the lower body joints.
Quantitatively, our disentangled strategy alone (\textbf{Ours w/o zoom}) obtains 6mm ($8\%$) and 5mm ($5.5\%$) improvement over \textbf{3DV'17} in the indoor and outdoor scenarios respectively. 
Using the features from the zoom-in branch in distance estimation (\textbf{Ours w/o averaging}) gains an additional improvement of 2mm and 1mm.
Using the averaged heatmaps (our full method) yields 7mm ($12.5\%$) and 6mm ($7.6\%$) improvement.
This evaluation shows that the 2D-3D consistency obtained by our disentangled strategy and the more accurate 2D prediction from the zoom-in branch are the key contributors to the overall improvement.

\subsection{Discussion}
We have demonstrated compelling real-time human 3D pose estimation results from a single cap-mounted fisheye camera.
Nevertheless our approach still has a few limitations that can be addressed in follow-up work:
1) Similar to all other learning-based approaches, it does not generalize well to data far outside the span of the training corpus.
This can be alleviated by extending the training corpus to cover larger variations in motion, body shape and appearance.
Since we train on synthetically rendered data, this is easily possible.
2) The reconstruction of 3D body pose under strong occlusions is challenging, since such situations are highly ambiguous, such as when the arms are raised above the head and thus cannot be seen by the camera.
In these cases, there are multiple distinct body poses that could give rise to the same observation, thus 3D pose estimation can fail.
Fortunately, since our approach works on a per-frame basis, it can recover directly after the occluded parts become visible again.
3) Our per-frame predictions may exhibit some temporal instability, similar to previous single-frame methods.
We believe that our approach could be easily extended by adding temporal stabilization as a post-process, or by using a recurrent architecture.
Several typical failure cases are shown in Fig.~\ref{fig:limitation}.
Despite these limitations, we believe, that we took an important step in the direction of real-time ubiquitous mobile 3D motion capture.
Our current capture setup conveniently augments a widely used fashion item. In future work, we will explore the design space more broadly and also experiment with other unconventional body-mounted camera locations.

\section{Applications}

Our egocentric 3D pose estimation system can be used for various applications such as action recognition, motion control, and performance analysis.
Especially, our system provides a novel natural human-computer-interaction (HCI) solution for recent popular virtual reality (VR) and augmented reality (AR) systems.
Specifically, body gesture-based HCI translates the natural movements of the user's body into tangible actions in a virtual world.
This allows the users to immerse themselves in a virtual environment and interact with the virtual content more intuitively.
Previous solutions typically rely on controllers or outside-in vision-based tracking systems.
In contrast, our system provides a compact, inside-in and controller-free solution, which can be integrated in the VR headsets and therefore does not require external tracking devices.
In VR games, our system allows the users to control a virtual character with their full body movements~\cite{Rhodin2015Generalizing}, instead of only hands for typical controller-based systems.
The users will also have a better perception of their full body movements, which is important typically for many sports games.
Similarly, the same technology can also be used for sports training or health-care, where the motion capture results can be used for motion analysis, performance monitoring or fall detection.
In VR/AR-based telepresence applications, our system can be used to capture the body motion, which can then be used to animate an avatar, without complicated multi-camera motion capture systems.
Importantly, benefiting from our mobile egocentric setup, users are not restricted to a fixed recording volume and therefore can roam freely while being captured.

\section{Conclusion}
\label{sec:conclusion}

We proposed the first real-time approach for 3D human pose estimation from a single fisheye camera that is attached to a standard baseball cap.
Our novel monocular setup clearly improves over cumbersome existing technologies and is an important step towards practical daily full-body motion capture.
3D pose estimation is based on a novel 3D pose regression network that is specifically tailored to our setup.
We train our network an a new ground truth training corpus of synthetic top-down fisheye images, which we will make publicly available.
Our evaluation shows that we achieve lower 3D joint error as well as better 2D overlay than exisiting baseline methods, when applied to the egocentric fisheye setting.
We see our approach as the basis for many exciting new applications in several areas, such as action recognition, performance analysis, and motion control in fields such as sports, health-care, and virtual reality.

%% if specified like this the section will be committed in review mode
\acknowledgments{
This work is supported by ERC Starting Grant “CapReal” (335545).
}

\bibliographystyle{abbrv-doi}

\bibliography{template}
\end{document}